\documentclass[conf]{IEEEtran}
\usepackage{algorithm,algpseudocode}
\usepackage{amsmath}
\usepackage{amssymb}
\usepackage{mathtools}
\usepackage[table]{xcolor}
\usepackage{bm}

\usepackage{multicol, multirow}
\usepackage{pifont}
\newcommand{\cmark}{\ding{51}}%
\newcommand{\xmark}{\ding{55}}%

\usepackage{cite}
\def\BibTeX{{\rm B\kern-.05em{\sc i\kern-.025em b}\kern-.08em
    T\kern-.1667em\lower.7ex\hbox{E}\kern-.125emX}}
\usepackage[bookmarksnumbered=true]{hyperref} 
\hypersetup{
     colorlinks = true,
     linkcolor = blue,
     anchorcolor = blue,
     citecolor = blue,
     filecolor = blue,
     urlcolor = blue
     }

\usepackage{svg}

\usepackage{wrapfig}
\usepackage{caption}
\usepackage{subcaption}
\usepackage{booktabs}

\usepackage{tikz}

\usepackage{xcolor}

\begin{document}

\captionsetup[table]{skip=0pt}
\captionsetup[figure]{skip=0pt}

\title{\textit{Are SNNs Truly Energy-efficient?} --- A Hardware Perspective}

\author{\IEEEauthorblockN{Abhiroop Bhattacharjee$^*$, Ruokai Yin$^*$, Abhishek Moitra$^*$, and
Priyadarshini Panda}\thanks{\rule{0.3\linewidth}{.5pt}\\$^*$Equal contribution. \\ This work was supported in part by CoCoSys, a JUMP2.0 center sponsored by DARPA and SRC, the NSF CAREER Award, TII (Abu Dhabi), and the DoE MMICC center SEA-CROGS (Award \#DE-SC0023198).}\\
\IEEEauthorblockA{Department of Electrical Engineering,
Yale University, USA\\
Email: \{abhiroop.bhattacharjee, ruokai.yin, abhishek.moitra, priya.panda\}@yale.edu} \vspace{-3ex}}




\maketitle

\begin{abstract}

Spiking Neural Networks (SNNs) have gained attention for their energy-efficient machine learning capabilities, utilizing bio-inspired activation functions and sparse binary spike-data representations. While recent SNN algorithmic advances achieve high accuracy on large-scale computer vision tasks, their energy-efficiency claims rely on certain impractical estimation metrics. This work studies two hardware benchmarking platforms for large-scale SNN inference, namely SATA and SpikeSim. SATA is a sparsity-aware systolic-array accelerator, while SpikeSim evaluates SNNs implemented on In-Memory Computing (IMC) based analog crossbars. Using these tools, we find that the actual energy-efficiency improvements of recent SNN algorithmic works differ significantly from their estimated values due to various hardware bottlenecks. We identify and addresses key roadblocks to efficient SNN deployment on hardware, including repeated computations \& data movements over timesteps, neuronal module overhead and vulnerability of SNNs towards crossbar non-idealities.

\end{abstract}

\begin{IEEEkeywords}
Spiking Neural Networks, Systolic-arrays, In-memory Computing, Crossbars, Energy-efficiency \vspace{-3mm}
\end{IEEEkeywords}

\IEEEpeerreviewmaketitle

\section{Introduction}
\label{sec:intro}
Spiking Neural Networks (SNNs) have garnered significant attention as a power-efficient solution for machine learning \cite{roy2019towards, yamazaki2022spiking}. SNNs process data over multiple time steps using biologically inspired non-linear activation functions, such as Leaky-integrate-and-Fire (LIF) neurons. During each time step, input data is represented as either a spike (binary 1) or no-spike (binary 0), creating a sparsely encoded temporal spike-data representation. This representation potentially offers several hardware advantages: (1) \textbf{Multiplier-less Computation:} SNNs use computation units that rely solely on accumulators for dot-products, eliminating the need for multipliers used in Artificial Neural Networks (ANNs) for Multiply-and-Accumulate (MAC) operations \cite{akopyan2015truenorth, tang2023hardware}. (2) \textbf{Reduced On-chip Memory:} The binary nature of SNNs significantly reduces the on-chip memory required to store intermediate layer activations during SNN processing. These features add to the energy-efficiency of SNN algorithms.



Fortunately, over the last few years, there have been huge advances in the SNN training algorithms \cite{lee2020enabling, chowdhury2021spatio, kim2021revisiting, wu2019direct, zhang2020temporal, zheng2021going, kim2022exploring, venkatesha2021federated, li2023seenn,li17efficient} leading to state-of-the-art classification accuracy at low timesteps on large-scale image datasets such as CIFAR10, CIFAR100,  TinyImageNet and ImageNet. However, being alogrithm-focused, the energy-efficiency claimed by these works are based on primitive metrics such as FLOPs, timesteps and spike-data sparsity. Such energy evaluation is impractical as metrics such as FLOPs do not account for hardware overheads like memory access and data communication. Additionally, real systolic-array \cite{xu2023survey} and In-memory Computing (IMC) \cite{verma2019memory} accelerators are ineffective in handling the spike-data sparsity, particularly during the memory fetches. Therefore, there is a need for realistic SNN hardware benchmarking platforms. As shown in Table \ref{tab:prior_works}, there are several SNN-specific hardware accelerators. Works such as BrainScale \cite{pehle2022brainscales}, Loihi \cite{davies2018loihi} and TrueNorth \cite{akopyan2015truenorth} are geared towards small-scale optimization tasks for SNNs. In more recent years, there have been several hardware co-design works \cite{narayanan2020spinalflow, lee2022parallel, ankit2017resparc, liang2021h2learn} that cater to large-scale SNN implementations. However, they lack several practical considerations such as the data communication overhead, LIF activation storage and hardware non-idealities \cite{bhattacharjee2022examining}.

\begin{table}[t]
\centering
\caption{Summary of Different Works}
\label{tab:prior_works}
\resizebox{0.9\columnwidth}{!}{\begin{tabular}{|c|c|c|c|}
\hline

\multirow{2}{*}{Work} & \multirow{2}{*}{\begin{tabular}{c} Training (\textbf{T}) or \\ Inference (\textbf{I}) \end{tabular}} & \multirow{2}{*}{\begin{tabular}{c} Platform \end{tabular}} & \multirow{2}{*}{\begin{tabular}{c} Hardware\\Benchmarking \end{tabular}}\\

& & & \\ \hline

\multicolumn{4}{|l|}{\cellcolor{blue!15} \textbf{Small-scale Optimization Tasks}} \\ \hline
BrainScale~\cite{van2012brain} & \textbf{T \& I} & Analog & \xmark \\ \hline
Loihi~\cite{davies2018loihi} & \textbf{I} & Digital & \xmark \\ \hline
TrueNorth~\cite{merolla2014million} & \textbf{I} & Digital & {\begin{tabular}{c} \xmark \end{tabular}} \\ \hline

\multicolumn{4}{|l|}{\cellcolor{blue!15} \textbf{Large-scale Computer Vision Tasks}} \\ \hline
{\begin{tabular}{c} SpinalFlow~\cite{narayanan2020spinalflow}, \\ PTB~\cite{lee2022parallel} \end{tabular}} 
 & \textbf{I} & Digital & {\xmark} \\ \hline


RESPARC~\cite{ankit2017resparc} & \textbf{I} & Analog & \xmark \\ \hline 

H2Learn~\cite{liang2021h2learn} & \textbf{T} & Digital & \xmark \\ \hline 

\cellcolor{green!15} SATA~\cite{yin2022sata} & \cellcolor{green!15} \textbf{T \& I} & \cellcolor{green!15} Digital & \cellcolor{green!15} \cmark \\ \hline 

\cellcolor{green!15} SpikeSim~\cite{moitra2023spikesim} & \cellcolor{green!15} \textbf{I} & \cellcolor{green!15} Analog & \cellcolor{green!15} \cmark \\ \hline 
\end{tabular} }
\vspace{-6mm}
\end{table}
To this end, we study two hardware accelerators SATA and SpikeSim. Unlike prior SNN hardware platforms \cite{narayanan2020spinalflow, lee2022parallel, ankit2017resparc, liang2021h2learn}, both SATA and SpikeSim support end-to-end hardware-realistic  benchmarking of large-scale SNNs, during inference. SATA \cite{yin2022sata} is a sparsity-aware systolic-array based training and inference accelerator for SNNs. 
While SATA evaluates SNN workloads on a fully-digital CMOS platform, SpikeSim \cite{moitra2023spikesim} performs hardware-realistic accuracy, energy, latency and area evaluation of SNN workloads on IMC analog crossbars based on Resistive Random-access Memories (RRAMs) \cite{chakraborty2020pathways}. 

Table \ref{tab:energy-comp} shows the estimated and hardware-realistic (SATA and SpikeSim implemented) energy-efficiency improvements of state-of-the-art SNN algorithms during inference. The estimated energy is proportional to the product of FLOPs, timesteps and the sparsity (as shown in footnote of Table \ref{tab:energy-comp}). Evidently, there is a significant difference between the estimated and hardware-realistic energy-efficiency improvements. 
To this end, in this work, we perform realistic SNN benchmarking on the SATA and SpikeSim platforms. With this, we bring forth the key bottlenecks that SNNs exhibit on hardware and propose effective mitigation strategies. Essentially, we address the following key bottlenecks: (1) repeated memory accesses and computations over multiple timesteps, (2) overhead of the LIF neuronal module, and (3) vulnerability of IMC-implemented SNNs towards analog crossbar non-idealities. This work encapsulates the above key hardware challenges overlooked by the SNN research community, and motivates future works aimed towards efficient hardware-aware SNN algorithm design.

\begin{table}[t]
    \centering 
    \caption{Table showing energy comparisons for recent SNN algorithm works using the CIFAR10 dataset. Est. Energy denotes the qualitative estimated energy (nJ) of the SNNs calculated using the equation specified in the footnote. $(N\times)$ denotes the estimated or actual improvements in energy as compared to the corresponding baselines in each work. $E_{AC}$ denotes the energy expended by a single INT8 Accumulation (AC) operation. All energy values are reported in 28 nm CMOS technology node. }
    \resizebox{\columnwidth}{!}{
    \begin{tabular}{ccccccc} \toprule

         \multirow{2}{*}{Work} & \multirow{2}{*}{Accuracy} & \multirow{2}{*}{Sparsity}  & \multirow{2}{*}{T} &  \multirow{1}{*}{Est. Energy$^2$} & \multicolumn{2}{c}{Actual Energy (nJ)} \\ 
         & & && (nJ) & SATA$^3$ & SpikeSim$^4$ \\ \midrule
         S-BP\cite{lee2020enabling}& 89.3\% & 90.0\%$^1$ & 50&$11.7e^4$($10\times$)& $1.1e^7$($4.5\times$) & $2.3e^5$(5.2$\times$) \\
         BNTT~\cite{kim2021revisiting}& 90.3\% & 90.5\% & 20&  $4.2e^4$(20$\times$) & $0.6e^7$(2.8$\times$) & $0.9e^5$(4$\times$) \\
        Direct~\cite{wu2019direct}& 90.5\% & 90.0\%$^1$ & 10&$11.7e^4$($20\times$)& $3.8e^7$($2.3\times$) & $2.6e^5$(4$\times$) \\TSSL~\cite{zhang2020temporal}& 91.4\% & 90.1\% & 5&$5.8e^4$(80$\times$) & $3.1e^7$($4.9\times$) & $1.3e^5$(16$\times$) \\
        LTH~\cite{kim2022exploring}& 93.2\% & 97.5\%$^1$ & 5&$2.7e^4$($15\times$)& $5.7e^7$($1.3\times$) & $2.6e^5$(2$\times$) \\
        TDBN~\cite{zheng2021going}& 92.9\% & 85.0\% & 4&$13.3e^4$(83$\times$)& $5.4e^7$($6.8\times$) & $2.1e^5$(25$\times$)  \\
         \bottomrule
         \multicolumn{7}{l}{$^1$ We use 90\% sparsity for~\cite{wu2019direct,lee2020enabling}. For~\cite{kim2022exploring}, we use the weight sparsity.}\\
         \multicolumn{7}{l}{$^2$$\bm{E}_{est} = FLOPs \times Timesteps \times (1-Sparsity) \times E_{AC}$}\\
         \multicolumn{7}{l}{$^3$ Codes are available at: \url{https://github.com/Intelligent-Computing-Lab-Yale/SATA}}\\
         \multicolumn{7}{l}{$^4$ Codes are available at: \url{https://github.com/Intelligent-Computing-Lab-Yale/SpikeSim}}\\
    \end{tabular}}
    \vspace{-6mm}
    \label{tab:energy-comp}
\end{table}
\vspace{-2mm}

\section{Background}

\noindent\textbf{Spiking Neural Networks:} The distinguishing feature of SNNs lies in their utilization of a different neuronal activation function (most commonly, LIF) for temporal signal processing, as opposed to the ReLU activation commonly used in ANNs. 
The LIF neuron $i$ associated with a membrane potential $u_{i}^{t}$, that accumulates a train of spike inputs as follows:
\vspace{-2mm}
\begin{equation}
    u_i^t = \lambda  u_i^{t-1} + \sum_j w_{ij}o^t_j.
    \label{eq:LIF}
\vspace{-3mm}
\end{equation}

Here, $t$ stands for the timestep, $w_{ij}$ for weight connections between neuron $i$ and neuron $j$ and $\lambda$ denotes the leak factor. The  LIF neuron $i$ generates an output spike $o_i^{t}$ at the end of timestep $t$ if the membrane potential exceeds a threshold $\theta$:
\vspace{-2mm}
\begin{equation}
    o^{t}_i =
\begin{cases}
 1,          & \text{if $u_i^{t} >\theta$},  \\
    0
    & \text{otherwise.} 
\end{cases}
\label{eq:firing}
\end{equation}
\vspace{-4mm}

Upon firing, the membrane potential is reset to zero.
The integrate-and-fire behavior exhibited by an LIF neuron results in a non-differentiable function, making it challenging to employ standard backpropagation for training SNNs. To this end, Surrogate gradient learning or Backpropagation Through Time (BPTT) addresses the non-differentiability problem of a LIF neuron by approximating the backward gradient function \cite{wu2018spatio} and offers a means to directly learn from spikes using fewer timesteps. Further, BPTT can be implemented using popular machine learning frameworks like PyTorch \cite{paszke2017automatic}.

Further, following previous work \cite{kim2022rate}, we use the direct encoding method to encode the input tensor into spike trains with total timesteps $T$. To get the final prediction, we repeat the inference process $T$ times ($t=1,2,...,T$) and average the output from the SNN output classifier.

\noindent\textbf{Systolic-array Accelerators:}  Systolic-array architecture is popular among the digital von-Neumann accelerator designs for SNNs~\cite{yin2022sata,narayanan2020spinalflow,lee2022parallel}. With a regular and dataflow centric design, systolic-arrays can efficiently process the matrix-matrix multiplications in parallel with high spatio-temporal locality
and compute density~\cite{kung1982systolic} (see Fig. \ref{fig:sata_arch}). In this work, we will evaluate the SATA design in two dataflow modes: 1) output-stationary (OS) mode, where the partial sums will remain inside each Processing Engine (PE) of the array during the dot-product operations; 2) weight-stationary (WS) mode, where, the weights are pre-stored into the PEs of the array before the dot-product operation starts.

\begin{wrapfigure}{l}{0.4\columnwidth}
 \centering
 \vspace{-4mm}
\includegraphics[width=0.4\columnwidth]{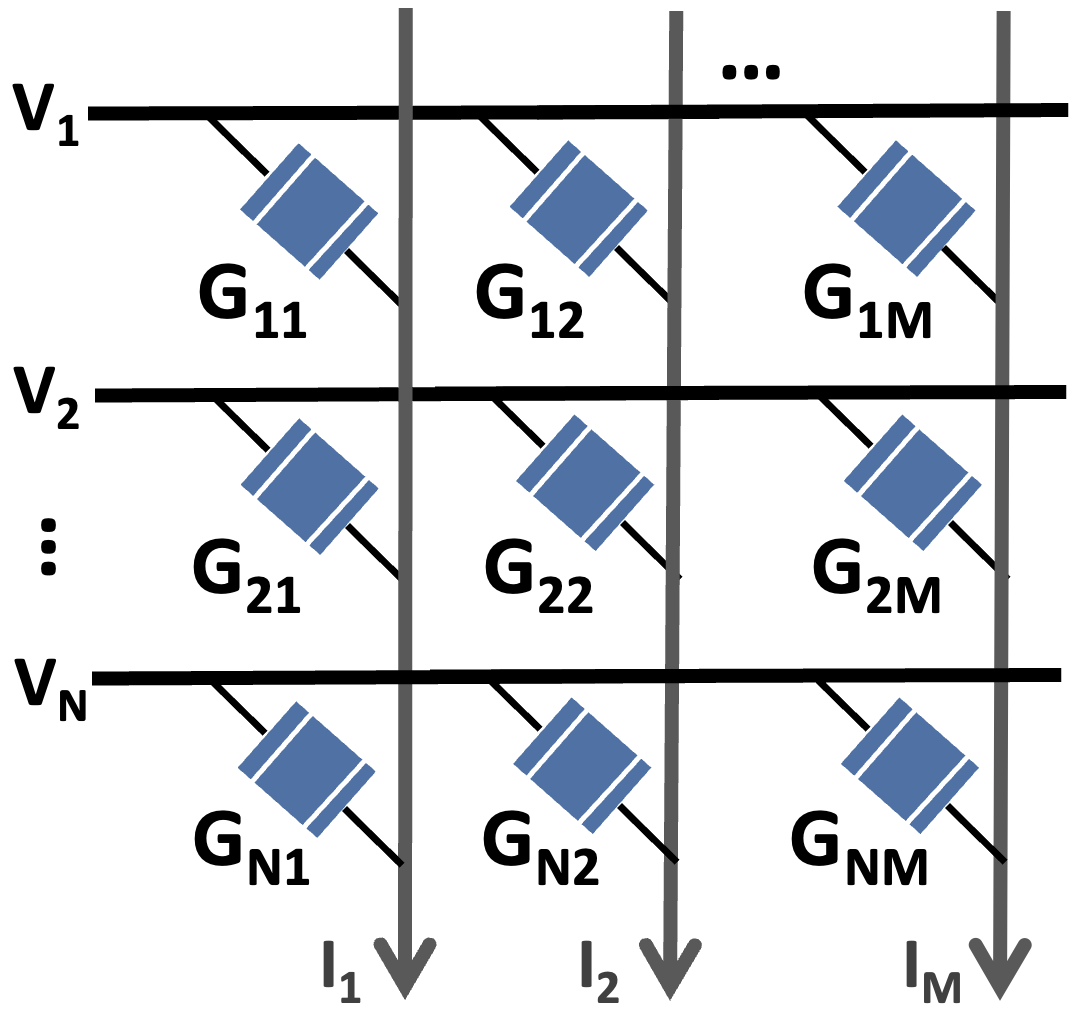}
\caption{An analog crossbar array with input voltages $V_{i}$, synaptic devices bearing conductances $G_{ij}$ and output currents $I_{j}$.}
\vspace{-3mm}
\label{xbar_img}
\end{wrapfigure}

\noindent\textbf{Analog Crossbars:} Analog crossbars comprise of a 2D array of IMC devices,  interfaced with Digital-to-Analog Converters (DACs), Analog-to-Digital Converters (ADCs), and write circuits dedicated towards programming the IMC devices \cite{hu2016dot, xia2019memristive, verma2019memory}. The SNN's spike inputs are encoded as analog voltages $V_i$ to each row of the crossbar by the DACs, while weights are programmed as synaptic device conductances ($G_{ij}$) at the cross-points, as shown in Fig. \ref{xbar_img}. For emulating dot-product operations in  case of an ideal N$\times$M crossbar during inference, the voltages interact with the device conductances, resulting in a current governed by Ohm's Law. Finally, adhering to Kirchoff's current law, the net output current sensed at each column $j$ by the ADCs represents the sum of currents flowing through each device, expressed as $I_{j(ideal)} = \Sigma_{i=1}^{N}{G_{ij} * V_i}$.  In practical scenarios, the analog nature of computation introduces various non-idealities, including interconnect parasitic resistances and synaptic device variations \cite{sun2019impact, bhattacharjee2021neat, jain2020rxnn, chakraborty2020geniex}. Thus in a non-ideal scenario, the net output current sensed at each column $j$ deviates from the ideal value $I_{j(ideal)}$. These deviations manifest as significant accuracy losses for SNNs on crossbars \cite{bhattacharjee2022examining}.

\section{Tools for Benchmarking SNNs on Hardware}

\subsection{SATA: A Systolic-array Benchmarking Accelerator }
\label{sec:sata}
\begin{figure}[t]
    \centering
\includegraphics[width=.8\linewidth]{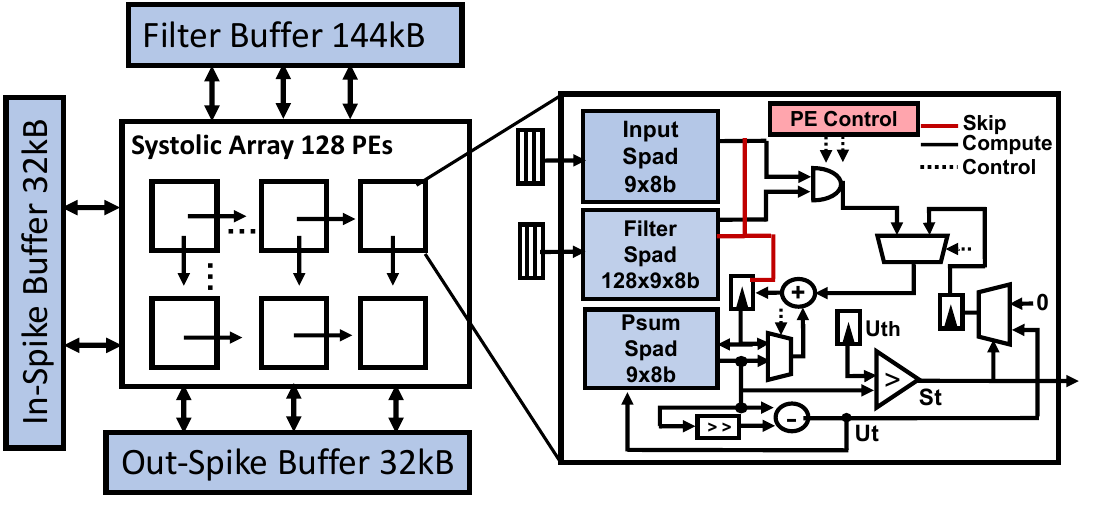}
    \caption{Architecture design of the inference version of the SATA tool. Spad refers to scratchpad memory (registers) and PE stands for Processing Engine that performs the weighted-accumulation and LIF operations. The red wire is used to indicate skipping of operations in case of zero operands as inputs.}
    \vspace{-4mm}
    \label{fig:sata_arch}
\end{figure}

SATA \cite{yin2022sata} is a sparsity-aware training accelerator designed for benchmarking the state-of-the-art BPTT-based SNN training on a fully digital von-Neumann architecture. Unlike the prior SNN training accelerators, which have numerous complex engines to boost performance, SATA adopts a simple and reconfigurable systolic-array design with a three-level memory hierarchy~\cite{chen2016eyeriss}. This makes it straightforward for the SNN algorithm designers to deploy their workloads on SATA and get an estimation of the hardware energy cost. Though designed as a training accelerator, SATA can be used to benchmark the inference performance of pre-trained SNNs by detaching the training-related components. This work will focus on using SATA to evaluate BPTT-trained SNNs during inference. The architecture design of SATA for inference is shown in Fig.~\ref{fig:sata_arch}. The analyses performed on SATA help identify some major bottlenecks, such as repetitive data movements across timesteps, that hinder SNNs from being energy-efficient on hardware. SATA shows that spike-data sparsity in SNNs can only be leveraged inside the PE computation unit (that performs weighted-accumulation and LIF operations). Outside the computation unit, even sparse input and weight data incur memory fetches from the on-chip buffers, adding to significant energy overhead. 

\noindent\textbf{Challenges:} Based on our benchmarking using SATA, we identify two key challenges for deploying SNNs on a systolic-array architecture. Firstly, increased timesteps bring in extra computation and data movement costs. As shown in Fig.~\ref{fig:acc_edp}(b), the energy costs for both PE computation and data movement from the on-chip buffers scales with increasing timesteps. These time-repetitive costs (specifically the data movement cost) reduce the energy-efficiency of the deployed SNNs. A prior work \cite{narayanan2020spinalflow} has shown that the inference energy gap between an SNN with 16 timesteps and its ANN counterpart can be as large as $16\times$ across various workloads on a systolic-array architecture, due to repetitive evaluations across multiple time-steps. The other challenge is attributed to the hardware cost of the LIF units which are used to generate the output spikes and store the membrane potential across timesteps.

\vspace{-4mm}

\subsection{SpikeSim: An IMC-based Benchamarking Accelerator}
\label{sec:spikesim}

\begin{figure}[t]
    \centering
    \includegraphics[width=.7\columnwidth]{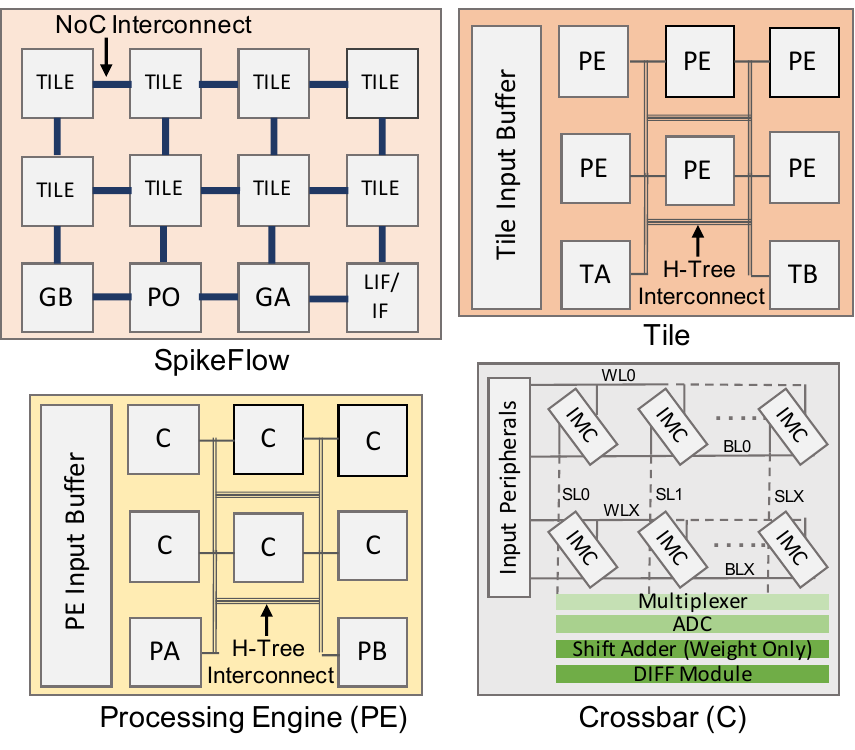}

    \caption{SpikeFlow Architecture. GB (GA), TB (TA) and PB (PA) denote global, tile and PE buffers (Accumulators). PO and LIF/IF denote the pooling and LIF/IF neuronal module, respectively. }
    \vspace{-6mm}
    \label{fig:spikeflow}
     
\end{figure}

SpikeSim \cite{moitra2023spikesim} is a IMC crossbar-based hardware evaluation tool for benchmarking SNN inference. SpikeSim maps BPTT-trained SNN workloads on a monolithic IMC-based weight-stationary tiled architecture, called SpikeFlow (see Fig. \ref{fig:spikeflow}), and performs hardware-realistic accuracy (incorporating crossbar non-idealities), energy, latency and area evaluations. SpikeFlow incorporates a digital Leaky-Integrate-Fire/Integrate-Fire (LIF/IF) neuronal activation unit to store the intermediate membrane potentials ($u^t$) and generate spike outputs during SNN inference. Furthermore, the analog crossbars in the SpikeFlow architecture are based on RRAM devices \cite{hajri2019rram}. For crossbar-realistic dot-product operations, SpikeSim emulates the impact of RRAM device read noise and IR-drop non-idealities due to crossbar interconnect parasitics. Another unique characteristic of SpikeSim is a fully digital DIFF module in the SpikeFlow architecture  that eliminates the conventional double-crossbar approach for performing signed dot-products, thereby bringing in energy and area savings. SpikeSim uses H-trees to communicate partial sums emerging from the crossbars to the digital peripherals inside a tile, and an inter-tile Network-on-Chip (NoC) architecture to communicate spikes to and from the neuronal unit. 

\noindent\textbf{Challenges:} Based on our benchmarking using SpikeSim, we bring forth three key challenges to SNN inference on IMC crossbar-based hardware. First, SNNs are highly vulnerable on analog crossbars owing to the impact of the non-idealities which lead to accumulation of errors in the dot-product operations over multiple timesteps (see Fig. \ref{fig:lif-ni-fig}(b)). Second, unlike ANNs that have a simple ReLU activation unit, SNNs entail a high LIF/IF neuronal area overhead on SpikeSim. This is due to large $u^t$ SRAM cache to store intermediate membrane potentials of different layers of an SNN model during inference. Third, unlike ANNs, the number of timesteps in SNNs plays a crucial role in the hardware performance and inference accuracy. As shown in Fig. \ref{fig:acc_edp}(a), both the inference energy and latency on SpikeSim scale linearly with timesteps, similar to the trend obtained using SATA in Fig. \ref{fig:acc_edp}(b).

\vspace{-3mm}

\section{Mitigation Strategies}
\label{sec:mitigation}
        

     

\noindent\textbf{Experimental Setup:} For all our experiments, we use BPTT-trained SNNs---VGG9, VGG16 and ResNet18 models, on the CIFAR10 and Tiny ImageNet datasets. We obtain SNN models using \href{https://github.com/Intelligent-Computing-Lab-Yale/BNTT-Batch-Normalization-Through-Time}{code} provided in \cite{kim2021revisiting}. Unless otherwise mentioned, all the pre-trained SNNs have 8-bit weight-precision. SATA \cite{yin2022sata} is calibrated in  28 nm CMOS technology node, while SpikeSim \cite{moitra2023spikesim} is in  65 nm CMOS technology node with RRAM crossbars of size 64$\times$64 in the PEs.

\begin{figure}[t]
\begin{center}
\def\arraystretch{0.5}
\begin{tabular}{@{}c@{}c}
\includegraphics[width=0.55\linewidth]{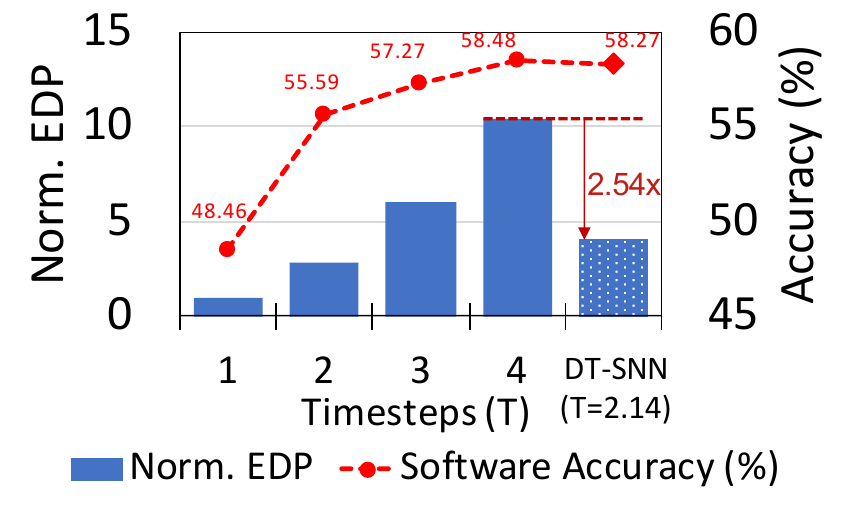} &
\includegraphics[width=0.45\linewidth]{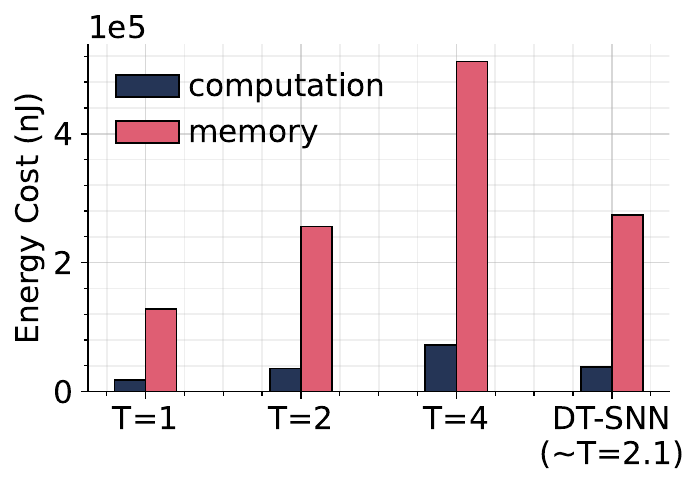} 
\\

{ (a)} & { (b)} 
\end{tabular}
\caption{(a) Impact of DT-SNN on SpikeSim using VGG16 SNN on the TinyImageNet dataset. The EDPs are normalized with respect to the value at timestep $= 1$. (b) Computation and memory movement costs on SATA across multiple timesteps for VGG9 SNN on the CIFAR10 dataset. Impact of DT-SNN on SATA energy costs.
}
\vspace{-7mm}
\label{fig:acc_edp}
\end{center}
\end{figure}

\noindent\textbf{Dynamic Timestep Reduction:} To improve the energy-efficiency of SNNs by reducing timesteps, we analyse an input-aware Dynamic Timestep SNN (DT-SNN) methodology \cite{li2023input}. DT-SNN dynamically determines the least number of timesteps required for a confident prediction in an input-dependent basis during inference of a pre-trained SNN. This is done by simply appending a digital entropy-computation module with our SNN-based hardware accelerators (SATA or SpikeSim). For every input, the calculated value of entropy of the SNN's predicted output at the end of every timestep is compared against a predefined threshold. An early temporal exit (termination of inference) or a prediction is carried out if the entropy is lower than the set threshold at any given timestep. Note, the entropy-computation module incurs negligible energy overhead to the overall SNN inference energy on SATA or SpikeSim. On SpikeSim, Fig. \ref{fig:acc_edp}(a) shows a $10.4\times$ higher Energy-Delay-Product (EDP) on increasing the number of timesteps from 1 to 4 for the inference of a standard VGG16 SNN on the Tiny ImageNet dataset. We find that DT-SNN can reduce the overall EDP by $2.54\times$, while maintaining similar inference accuracy, compared to a standard SNN inference with 4 timesteps across all inputs. It turns out a large fraction of the test images in the Tiny ImageNet dataset are classified with 1 timestep, and overall DT-SNN requires 2.14 timesteps on average across all test images resulting in lower EDP. On SATA, a similar trend of inference energy increase is observed for both computation and memory access energy when the timestep increases from 1 to 4. With DT-SNN, the total energy cost is reduced by $46.5\%$ compared to the standard 4-timestep VGG9 SNN inference on CIFAR10 dataset.

\begin{figure}[t]
\begin{center}
\def\arraystretch{0.5}
\begin{tabular}{@{}c@{}c}
\includegraphics[width=0.5\linewidth]{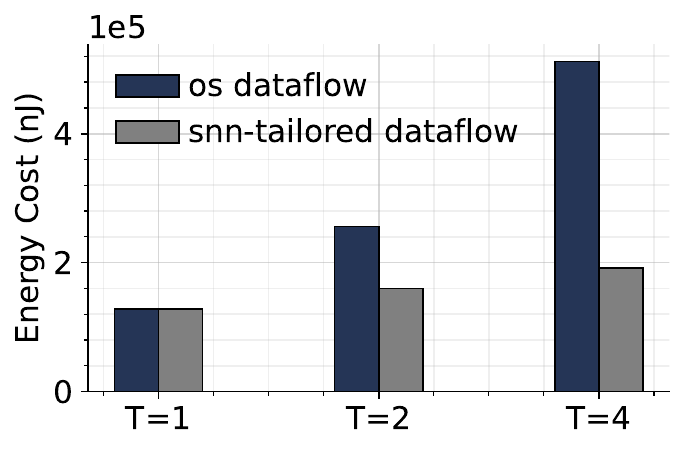} &
\includegraphics[width=0.5\linewidth]{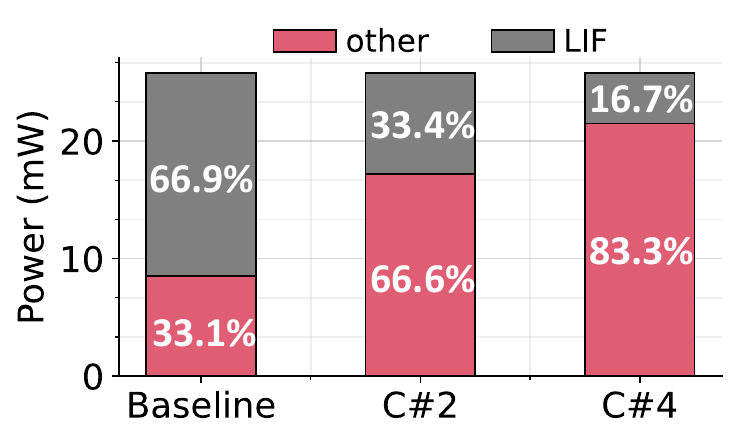} 
\\
{ (a)} & { (b)} 
\end{tabular}
\caption{(a) Data movement energy cost difference between a standard OS dataflow and the SNN-tailored dataflow for SATA using VGG9 SNN on Tiny ImageNet dataset. (b) The power cost breakdown of computation units on SATA with 128 PEs. The baseline has 128 LIF units. To deploy $C\#n$ EfficientLIF-Net, SATA requires 1/n of the baseline LIF units.
}
\vspace{-7mm}
\label{fig:sata-results}
\end{center}
\end{figure}

\noindent\textbf{Data-movement Cost Mitigation:} As we discussed in Section \ref{sec:sata}, one of the major challenges for SNN deployment on digital hardware platforms like SATA is the repetitive data movement costs. 
We design an SNN-tailored dataflow for SATA that can significantly reduce the repetitive data  movement cost for SNNs. In SATA's dataflow design, we adopt the tick-batch method~\cite{narayanan2020spinalflow} and maximally reuse the weights at the PE level by having scratch-pad memories inside every PE to hold the weights stationary~\cite{chen2016eyeriss} throughout the PE computation process. By utilizing such a dataflow, SATA will only read out the data once from the higher memory hierarchies (DRAM and SRAM) to the PE array for each layer across all timesteps. In Fig.~\ref{fig:sata-results}~(a), we compare SATA's SNN-tailored dataflow with the standard output-stationary dataflow on a VGG9 SNN on the Tiny ImageNet dataset with different timesteps. Utilizing the SNN-tailored dataflow can save $62.5\%$ memory movement energy with a timestep of 4. The benefits will increase when a larger timestep is used. Besides re-designing the dataflow for the hardware, model compression techniques like quantization~\cite{yin2023mint,tan2023low,deng2021comprehensive} and pruning\cite{yin2023workload,kim2022exploring,deng2021comprehensive} will also help in reducing the data movement costs.

\begin{figure}[h]
\begin{center}
\def\arraystretch{0.5}
\begin{tabular}{@{}c@{}c}
\includegraphics[width=0.45\linewidth]{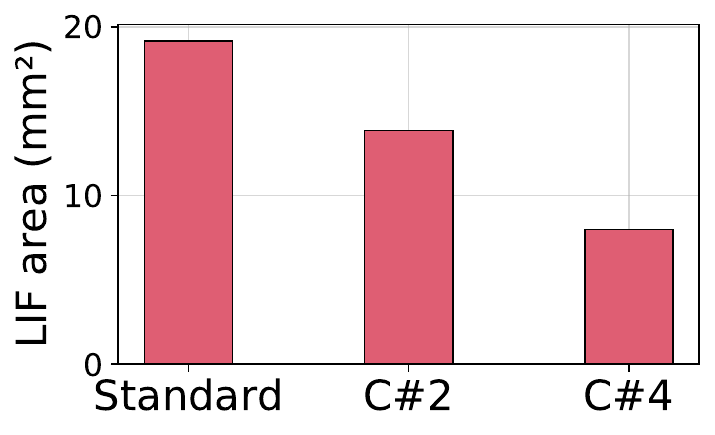} &
\includegraphics[width=0.55\linewidth]{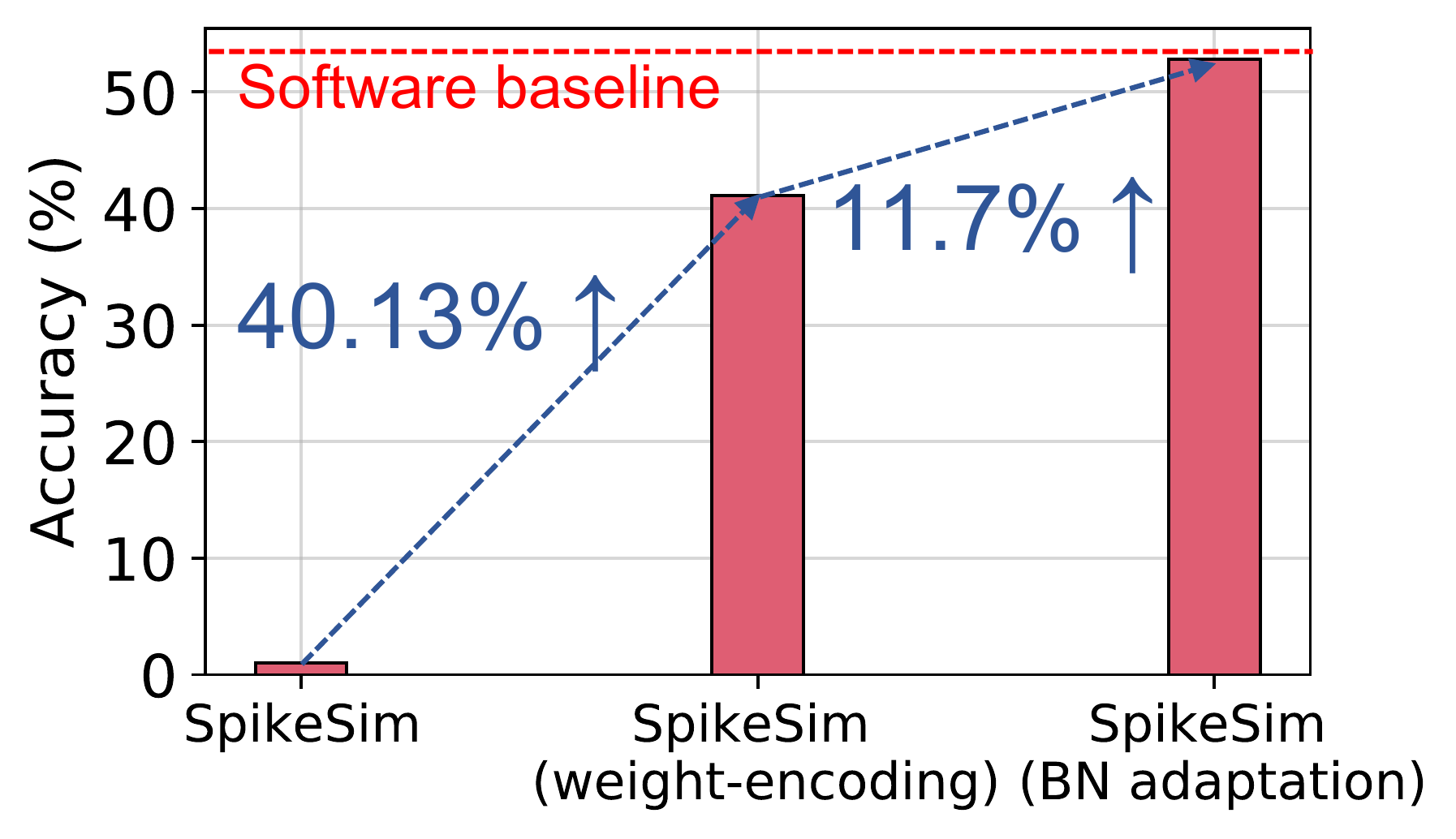} 
\\
{ (a)} & { (b)} 
\end{tabular}
\caption{(a) Impact of LIF-sharing on neuronal area using SpikeSim for ResNet18 SNN on Tiny ImageNet dataset. (b) Non-ideal accuracy improvement on SpikeSim for a pre-trained VGG16 SNN (4-bit weights) on Tiny ImageNet dataset with non-ideality-aware weight-encoding \& BN adaptation.
}
\vspace{-8mm}
\label{fig:lif-ni-fig}
\end{center}
\end{figure}
\noindent\textbf{Mitigating LIF Overhead:} LIF units are energy-hungry components on the hardware, which can take up to 61.6\% total power of the computation units~\cite{kim2023sharing}. This translates to approximately $2\times$ higher energy cost for LIF operations compared to other operations. To mitigate the overhead of LIF units, recent work EfficientLIF-Net~\cite{kim2023sharing} shares the LIF neurons across layers and channels. We use the notation of $C\#n$ to represent the EfficientLIF-Nets that share 1 LIF neuron for $n$ post-synaptic neurons on the output channel dimension.
On SATA, as shown in Fig.~\ref{fig:sata-results}~(b), we can reduce 75.1\% of the power cost of LIF units by having a $C\#4$ LIF-sharing. Quantization of membrane potential~\cite{yin2023mint} can also help to reduce the LIF unit cost by having smaller registers for the membrane potentials. The LIF-sharing and membrane potential quantization methods are orthogonal techniques for mitigating the LIF-units cost. On SpikeSim, we find that LIF sharing with $C\#2$ and $C\#4$ results in $1.38\times$ and $2.41\times$ reductions in LIF area, respectively, for a ResNet18 SNN on the TinyImagenet dataset (see Fig.~\ref{fig:lif-ni-fig}(a)). 

\noindent\textbf{Crossbar Non-ideality Mitigation:} To address the accuracy degradation of SNNs on non-ideal crossbars, two training-less approaches are studied in Fig. \ref{fig:lif-ni-fig}(b): (1) SpikeSim supports a non-ideality aware weight-encoding scheme on the RRAM crossbars to increase the proportion of high resistance synapses during SNN inference. Prior works have shown that the impact of crossbar non-idealities decreases upon increasing the proportion of high resistance synapses in the crossbars \cite{bhattacharjee2021neat, bhattacharjee2020switchx, bhattacharjee2022examining1}. Thus, the non-ideal SNN accuracy is improved by $40.13\%$. (2) Non-ideality aware adaptation \cite{bhattacharjee2022examining, bhattacharjee2023examining} of the SNN's batchnorm (BN) parameters prior to inference can mitigate the impact of crossbar non-idealities, specifically the interconnect parasitics. During BN adaptation, we forward a number of training image samples through the SNN deployed on crossbars, adapting the moving average \& variance of the batchnorm layers with respect to noisy activations (while keeping the learnable parameters or weights frozen). Consequently, the measured accuracy loss on SpikeSim due to non-idealities is reduced to $1.22\%$ compared to the software baseline. 

\vspace{-2mm}





\section{Conclusion}

\label{sec:conclusion}

Our study unveils critical challenges in efficient deployment of large-scale SNNs on hardware, highlighting discrepancies between estimated and hardware-realistic energy-efficiency improvements. We base our study using two state-of-the-art hardware benchmarking tools for SNNs (SATA and SpikeSim) which help identify and address key hardware bottlenecks such as- repeated computations \& data movements over timesteps, LIF neuronal module overhead and SNN's vulnerability to crossbar non-idealities. We identify some key mitigation strategies that help address the hardware overheads. The findings from the benchmarking tools underscore the importance realistic hardware-aware SNN algorithm-design in the future for driving low-power neuromorphic applications at the edge.



\scriptsize
\bibliographystyle{IEEEtran}
\bibliography{reference}

\begin{thebibliography}{10}
\providecommand{\url}[1]{#1}
\csname url@samestyle\endcsname
\providecommand{\newblock}{\relax}
\providecommand{\bibinfo}[2]{#2}
\providecommand{\BIBentrySTDinterwordspacing}{\spaceskip=0pt\relax}
\providecommand{\BIBentryALTinterwordstretchfactor}{4}
\providecommand{\BIBentryALTinterwordspacing}{\spaceskip=\fontdimen2\font plus
\BIBentryALTinterwordstretchfactor\fontdimen3\font minus
  \fontdimen4\font\relax}
\providecommand{\BIBforeignlanguage}[2]{{%
\expandafter\ifx\csname l@#1\endcsname\relax
\typeout{** WARNING: IEEEtran.bst: No hyphenation pattern has been}%
\typeout{** loaded for the language `#1'. Using the pattern for}%
\typeout{** the default language instead.}%
\else
\language=\csname l@#1\endcsname
\fi
#2}}
\providecommand{\BIBdecl}{\relax}
\BIBdecl

\bibitem{roy2019towards}
K.~Roy \emph{et~al.}, ``Towards spike-based machine intelligence with
  neuromorphic computing,'' \emph{Nature}, 2019.

\bibitem{yamazaki2022spiking}
K.~Yamazaki \emph{et~al.}, ``Spiking neural networks and their applications: A
  review,'' \emph{Brain Sciences}, 2022.

\bibitem{akopyan2015truenorth}
F.~Akopyan \emph{et~al.}, ``Truenorth: Design and tool flow of a 65 mw 1
  million neuron programmable neurosynaptic chip,'' \emph{IEEE transactions on
  computer-aided design of integrated circuits and systems}, 2015.

\bibitem{tang2023hardware}
C.~Tang and J.~Han, ``Hardware efficient weight-binarized spiking neural
  networks,'' in \emph{2023 Design, Automation \& Test in Europe Conference \&
  Exhibition (DATE)}.\hskip 1em plus 0.5em minus 0.4em\relax IEEE, 2023.

\bibitem{lee2020enabling}
C.~Lee \emph{et~al.}, ``Enabling spike-based backpropagation for training deep
  neural network architectures,'' \emph{Frontiers in neuroscience}, p. 119,
  2020.

\bibitem{chowdhury2021spatio}
S.~S. Chowdhury \emph{et~al.}, ``Spatio-temporal pruning and quantization for
  low-latency spiking neural networks,'' in \emph{2021 International Joint
  Conference on Neural Networks (IJCNN)}.\hskip 1em plus 0.5em minus
  0.4em\relax IEEE, 2021.

\bibitem{kim2021revisiting}
Y.~Kim and P.~Panda, ``Revisiting batch normalization for training low-latency
  deep spiking neural networks from scratch,'' \emph{Frontiers in
  neuroscience}, 2021.

\bibitem{wu2019direct}
Y.~Wu \emph{et~al.}, ``Direct training for spiking neural networks: Faster,
  larger, better,'' in \emph{Proceedings of the AAAI conference on artificial
  intelligence}, 2019.

\bibitem{zhang2020temporal}
W.~Zhang and P.~Li, ``Temporal spike sequence learning via backpropagation for
  deep spiking neural networks,'' \emph{Advances in Neural Information
  Processing Systems}, vol.~33, pp. 12\,022--12\,033, 2020.

\bibitem{zheng2021going}
H.~Zheng \emph{et~al.}, ``Going deeper with directly-trained larger spiking
  neural networks,'' in \emph{Proceedings of the AAAI conference on artificial
  intelligence}, 2021.

\bibitem{kim2022exploring}
Y.~Kim \emph{et~al.}, ``Exploring lottery ticket hypothesis in spiking neural
  networks,'' in \emph{European Conference on Computer Vision}.\hskip 1em plus
  0.5em minus 0.4em\relax Springer, 2022.

\bibitem{venkatesha2021federated}
Y.~Venkatesha \emph{et~al.}, ``Federated learning with spiking neural
  networks,'' \emph{IEEE Transactions on Signal Processing}, 2021.

\bibitem{li2023seenn}
Y.~Li \emph{et~al.}, ``Seenn: Towards temporal spiking early-exit neural
  networks,'' \emph{arXiv:2304.01230}, 2023.

\bibitem{li17efficient}
Li \emph{et~al.}, ``Efficient human activity recognition with spatio-temporal
  spiking neural networks,'' \emph{Frontiers in Neuroscience}, vol.~17, p.
  1233037.

\bibitem{xu2023survey}
R.~Xu \emph{et~al.}, ``A survey of design and optimization for systolic array
  based dnn accelerators,'' \emph{ACM Computing Surveys}, 2023.

\bibitem{verma2019memory}
N.~Verma \emph{et~al.}, ``In-memory computing: Advances and prospects,''
  \emph{IEEE Solid-State Circuits Magazine}, 2019.

\bibitem{pehle2022brainscales}
C.~Pehle \emph{et~al.}, ``The brainscales-2 accelerated neuromorphic system
  with hybrid plasticity,'' \emph{Frontiers in Neuroscience}, 2022.

\bibitem{davies2018loihi}
M.~Davies \emph{et~al.}, ``Loihi: A neuromorphic manycore processor with
  on-chip learning,'' \emph{Ieee Micro}, 2018.

\bibitem{narayanan2020spinalflow}
S.~Narayanan \emph{et~al.}, ``Spinalflow: An architecture and dataflow tailored
  for spiking neural networks,'' in \emph{International Symposium on Computer
  Architecture (ISCA)}.\hskip 1em plus 0.5em minus 0.4em\relax IEEE, 2020.

\bibitem{lee2022parallel}
J.-J. Lee \emph{et~al.}, ``Parallel time batching: Systolic-array acceleration
  of sparse spiking neural computation,'' in \emph{International Symposium on
  High-Performance Computer Architecture (HPCA)}.\hskip 1em plus 0.5em minus
  0.4em\relax IEEE, 2022.

\bibitem{ankit2017resparc}
A.~Ankit \emph{et~al.}, ``Resparc: A reconfigurable and energy-efficient
  architecture with memristive crossbars for deep spiking neural networks,'' in
  \emph{Design Automation Conference}, 2017.

\bibitem{liang2021h2learn}
L.~Liang \emph{et~al.}, ``H2learn: High-efficiency learning accelerator for
  high-accuracy spiking neural networks,'' \emph{IEEE Transactions on
  Computer-Aided Design of Integrated Circuits and Systems}, 2021.

\bibitem{bhattacharjee2022examining}
A.~Bhattacharjee \emph{et~al.}, ``Examining the robustness of spiking neural
  networks on non-ideal memristive crossbars,'' \emph{International Symposium
  on Low Power Electronics and Design (ISLPED)}, 2022.

\bibitem{van2012brain}
I.~L. van Soelen \emph{et~al.}, ``Brain scale: brain structure and cognition:
  an adolescent longitudinal twin study into the genetic etiology of individual
  differences,'' \emph{Twin Research and Human Genetics}, 2012.

\bibitem{merolla2014million}
P.~A. Merolla \emph{et~al.}, ``A million spiking-neuron integrated circuit with
  a scalable communication network and interface,'' \emph{Science}, 2014.

\bibitem{yin2022sata}
R.~Yin \emph{et~al.}, ``Sata: Sparsity-aware training accelerator for spiking
  neural networks,'' \emph{IEEE Transactions on Computer-Aided Design of
  Integrated Circuits and Systems}, 2022.

\bibitem{moitra2023spikesim}
A.~Moitra \emph{et~al.}, ``Spikesim: An end-to-end compute-in-memory hardware
  evaluation tool for benchmarking spiking neural networks,'' \emph{IEEE TCAD},
  2023.

\bibitem{chakraborty2020pathways}
I.~Chakraborty \emph{et~al.}, ``Pathways to efficient neuromorphic computing
  with non-volatile memory technologies,'' \emph{Applied Physics Reviews},
  2020.

\bibitem{wu2018spatio}
Y.~Wu \emph{et~al.}, ``Spatio-temporal backpropagation for training
  high-performance spiking neural networks,'' \emph{Frontiers in neuroscience},
  2018.

\bibitem{paszke2017automatic}
A.~Paszke \emph{et~al.}, ``Automatic differentiation in pytorch,'' 2017.

\bibitem{kim2022rate}
Y.~Kim \emph{et~al.}, ``Rate coding or direct coding: Which one is better for
  accurate, robust, and energy-efficient spiking neural networks?'' in
  \emph{ICASSP}.\hskip 1em plus 0.5em minus 0.4em\relax IEEE, 2022.

\bibitem{kung1982systolic}
H.-T. Kung, ``Why systolic architectures?'' \emph{Computer}, 1982.

\bibitem{hu2016dot}
M.~Hu \emph{et~al.}, ``Dot-product engine for neuromorphic computing:
  Programming 1t1m crossbar to accelerate matrix-vector multiplication,'' in
  \emph{ACM/EDAC/IEEE DAC}, 2016.

\bibitem{xia2019memristive}
Q.~Xia and J.~J. Yang, ``Memristive crossbar arrays for brain-inspired
  computing,'' \emph{Nature materials}, 2019.

\bibitem{sun2019impact}
Sun \emph{et~al.}, ``Impact of non-ideal characteristics of resistive synaptic
  devices on implementing convolutional neural networks,'' \emph{IEEE JETCAS},
  2019.

\bibitem{bhattacharjee2021neat}
A.~Bhattacharjee \emph{et~al.}, ``Neat: Non-linearity aware training for
  accurate, energy-efficient and robust implementation of neural networks on
  1t-1r crossbars,'' \emph{IEEE TCAD}, 2021.

\bibitem{jain2020rxnn}
S.~Jain \emph{et~al.}, ``Rxnn: A framework for evaluating deep neural networks
  on resistive crossbars,'' \emph{IEEE TCAD}, 2020.

\bibitem{chakraborty2020geniex}
I.~Chakraborty \emph{et~al.}, ``Geniex: A generalized approach to emulating
  non-ideality in memristive xbars using neural networks,'' in \emph{ACM/IEEE
  DAC}, 2020.

\bibitem{chen2016eyeriss}
Y.-H. Chen \emph{et~al.}, ``Eyeriss: A spatial architecture for
  energy-efficient dataflow for convolutional neural networks,'' \emph{ACM
  SIGARCH Computer Architecture News}, 2016.

\bibitem{hajri2019rram}
B.~Hajri \emph{et~al.}, ``Rram device models: A comparative analysis with
  experimental validation,'' \emph{IEEE Access}, 2019.

\bibitem{li2023input}
Y.~Li \emph{et~al.}, ``Input-aware dynamic timestep spiking neural networks for
  efficient in-memory computing,'' \emph{Design and Automation Conference},
  2023.

\bibitem{yin2023mint}
R.~Yin \emph{et~al.}, ``Mint: Multiplier-less integer quantization for spiking
  neural networks,'' \emph{arXiv:2305.09850}, 2023.

\bibitem{tan2023low}
P.-Y. Tan and C.-W. Wu, ``A low-bitwidth integer-stbp algorithm for efficient
  training and inference of spiking neural networks,'' in \emph{Proceedings of
  the 28th Asia and South Pacific Design Automation Conference}, 2023, pp.
  651--656.

\bibitem{deng2021comprehensive}
L.~Deng \emph{et~al.}, ``Comprehensive snn compression using admm optimization
  and activity regularization,'' \emph{IEEE transactions on neural networks and
  learning systems}, 2021.

\bibitem{yin2023workload}
R.~Yin \emph{et~al.}, ``Workload-balanced pruning for sparse spiking neural
  networks,'' \emph{arXiv:2302.06746}, 2023.

\bibitem{kim2023sharing}
Y.~Kim \emph{et~al.}, ``Sharing leaky-integrate-and-fire neurons for
  memory-efficient spiking neural networks,'' \emph{arXiv:2305.18360}, 2023.

\bibitem{bhattacharjee2020switchx}
A.~Bhattacharjee and P.~Panda, ``Switchx: Gmin-gmax switching for
  energy-efficient and robust implementation of binarized neural networks on
  reram xbars,'' \emph{ACM TODAES}, 2021.

\bibitem{bhattacharjee2022examining1}
Bhattacharjee \emph{et~al.}, ``Examining and mitigating the impact of crossbar
  non-idealities for accurate implementation of sparse deep neural networks,''
  in \emph{Design, Automation \& Test in Europe Conference \& Exhibition
  (DATE)}.\hskip 1em plus 0.5em minus 0.4em\relax IEEE, 2022.

\bibitem{bhattacharjee2023examining}
A.~Bhattacharjee \emph{et~al.}, ``Examining the role and limits of batchnorm
  optimization to mitigate diverse hardware-noise in in-memory computing,''
  \emph{GLSVLSI}, 2023.

\end{thebibliography}
\normalsize

\end{document}